IJSAB
International

# Plant Leaf Disease Detection and Classification Using Deep Learning: A Review and A Proposed System on Bangladesh's Perspective

**Md. Jalal Uddin Chowdhury, Zumana Islam Mou, Rezwana Afrin & Shafkat Kibria**

**Abstract**

A very crucial part of Bangladeshi people's employment, GDP contribution, and mainly livelihood is agriculture. It plays a vital role in decreasing poverty and ensuring food security. Plant diseases are a serious stumbling block in agricultural production in Bangladesh. At times, humans can't detect the disease from an infected leaf with the naked eye. Using inorganic chemicals or pesticides in plants when it's too late leads in vain most of the time, deposing all the previous labor. The deep-learning technique of leaf-based image classification, which has shown impressive results, can make the work of recognizing and classifying all diseases trouble-less and more precise. In this paper, we've mainly proposed a better model for the detection of leaf diseases. Our proposed paper includes the collection of data on three different kinds of crops: bell peppers, tomatoes, and potatoes. For training and testing the proposed CNN model, the plant leaf disease dataset collected from Kaggle, is used which has 17430 images. The images are labeled with 14 separate classes of damage. The developed CNN model performs efficiently and could successfully detect and classify the tested diseases. The proposed CNN model may have great potency in crop disease management.

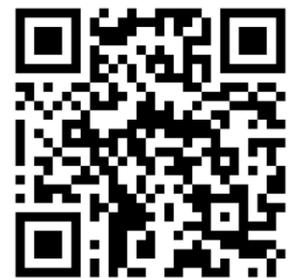

About Author (s)

**Md. Jalal Uddin Chowdhury** (corresponding author)**,** Department of CSE, Leading University, Sylhet, Bangladesh.
**Zumana Islam Mou,** Department of CSE, Leading University, Sylhet, Bangladesh.
**Rezwana Afrin,** Department of CSE, Leading University, Sylhet. Bangladesh.
**Shafkat Kibria**, Department of CSE, Leading University, Sylhet, Bangladesh.

**Introduction**

Bangladesh is a country which is known for its abundant fertile plain land. As a result of that the sustenance of its population and the growth of its economy heavily relies on agriculture. Agriculture plays a pivotal role in employing a significant portion of this country's population. Taking into consideration its fertile lands, diverse climatic conditions, and abundant water resources, Bangladesh possesses remarkable agricultural potential. Agriculture serves as the backbone of Bangladesh's economy as it contributes to its GDP to a great extent (Rahman, 2017). The food security and overall development of this nation go hand in hand with the development of this sector. Farmers worldwide have always been significantly and notably concerned regarding plant diseases and Bangladesh is no exception. The impact of plant diseases can play a crucial role in the degradation of a country's economy, especially in a country whose economy prominently depends on agriculture. The consequences of plant diseases are severe and their impact is particularly pronounced in staple crops like rice, potato, tomato, pepper, fruits, etc (Akhter et al., 2019). which form the core of the nation's food security? The country's farmers strive to meet the growing demands for food, feed, fiber, and bioenergy, all while dealing with numerous factors that exacerbate the vulnerability of crops to diseases. When crops succumb to diseases, the income of farmers gets reduced but at the same time, the production costs elevate due to excessive pesticide use and diminished access to markets due to the deterioration of produce quality (Rahman, 2017). Hence it is imperative to detect crop diseases at their early stage. Nonetheless, the farmers of Bangladesh are not equipped with remarkably advanced agricultural technologies that can detect the disease if the images are provided. They mostly try to detect the diseases manually with their naked eyes which a lot of time cannot identify accurately and is also very time consuming. So, for the early detection of affected plants, Bangladesh should adapt to modern technological systems rather than manual detection. Bangladesh is a South Asian riverine country. Its geographical location allows it to have fertile lands where numerous crops can be produced. Among these abundant crops rice, potato, tomato, and pepper can be considered as vital crops owing to the fact that these play an indispensable role in the growth of our country's entire economy, food security, and environmental sustainability (Islam et al., 2019). Yet these crops are vulnerable to various diseases which pose a great threat to agricultural productivity and the livelihood of farmers. In Bangladesh, rice production provides for half of the GDP that comes from the agricultural sector. Rice plays a pivotal role as it acts as the main food for the mass population of this country as well as dominating its economy (Ahmed , 2004). Being the staple food of Bangladesh, rice is prone to various diseases such as brown spot, bacterial blight, stem rot, rice blast, etc. Back then rice brown and blast spot were regarded as the most eminent disease but now brown spot and bacterial blight are considered as the most significant and menacing rice diseases (Al-Amin et al., 2019). Tomato holds immense significance due to its nutritional value and wide use in the culinary of Bangladesh (Samshunnahar et al. ,2016). Despite having plenty of virtues, it also could not escape from the wrath of various diseases. It includes leaf mold, TYLCV, late blight, Septoria leaf spot, early blight, etc (Tipu et al. ,2021). These occur due to the attack of bacteria or fungi. Potato is considered one of the most versatile plants since it can be consumed in multiple ways. It is a reliable source of carbohydrates, minerals, and vitamins. But because of some diseases such as early blight, late blight, etc. The production of potato gets hampered significantly. Pepper plays an important role in the growth of the agricultural sector of Bangladesh as it is a key ingredient in local dishes. In addition, pepper has potential health benefits as it is rich in antioxidants, vitamins, and capsaicin (Islam et al. ,2017). However, the cultivation of pepper is no exception as it is also prone to some diseases such as bacterial spot, phytophthora blight, etc. These diseases can impact the production and quality of the plants negatively. In this paper, we tried to shed light on a potential pathway for plant disease detection. A CNN model was proposed so that the diseases of three crops: bell pepper, tomato, and potato can be detected with the help of the images of the affected crops. The dataset of the crops' images was obtained from Kaggle (Plant Village., 2020). This dataset consists of a total of 17,430 images which were further divided into training and testing datasets. Further, this model was compared to other algorithms of machine learning such as KNN, logistic regression, SVM, and decision tree. However, our proposed model performed the best in correctly classifying and identifying the diseases and gave the most accuracy.

The purpose of this paper is to review different kinds of plant leaf detection systems using different types of machine learning algorithms which will facilitate the early detection of diseases in plants. As a result of that the farmers can be benefited immensely since early detection of plant leaves diseases will help to control the damage in a more effective manner and it will definitely give a boost to the production of crops. Besides, people who are interested in this field and are new to this research field would also be profited by this paper because by reviewing plenty of papers we tried to summarize all these and tried to find out the most efficient machine learning algorithms. According to the findings from the papers that were reviewed a CNN model was proposed for plant disease detection. A dataset of a total of 17,430 images was collected from Kaggle (Tipu et al., 2021). Further, the images were processed and divided into training and testing datasets. We compared our proposed CNN model to other machine learning algorithms and discovered that our model performs the best with the most accuracy.

**Literature Review**
A great deal of thought has been accorded to the problem of how to track and detect the spread of diseases that have a significant effect on plant leaves among the research community. Fenu and Malloci (2019) have investigated whether weather data from regional meteorological stations could be used to predict a risk index for potato late blight which was focused on the development of a feed-forward neural network and a support vector machine classification. The historical weather data was provided by ARPAS Agency from the Cagliari location and based on these data, a new dataset was built using the mathematical model SimCast. The SVM classification model performed better than the ANN model where the prediction accuracy for ANN was 96% and for SVM Classification was 98%. Ahmad et al. (2023) analyzed almost seventy studies on deep learning applications for plant disease classification and management in agriculture. The review focused on key research questions related to deep learning techniques, dataset requirements, imaging sensors, disease severity estimation, generalization of models, comparison with human accuracy, and open research topics. The studies covered various aspects of plant disease diagnosis, including machine learning, image processing, and deep learning. The review also went through the performance of deep learning models and their potential to perform better than humans. The findings can contribute to an automated plant disease management system by enabling early identification and accurate severity estimation. Ramesh et al. (2018) conducted plant disease detection using the machine learning algorithm. They trained Six models using their own dataset for improved detection. These algorithms include Logistic Regression, Random Forest Classification, K-Nearest Neighbors, Support vector machine, CART, and Naive Bayes Classification. The Random forest

Classification achieved the best results, with an accuracy rate of 70%. Nevertheless, this investigation was restricted to a small number of datasets, specifically 160 datasets for training and testing. So its findings should be interpreted with caution.

Bashish et al. (2010) proposed a framework that is an image-processing-based approach used for leaf and stem disease detection. The proposed framework relies on the following stages of image processing. K-means is used to first separate out certain regions within an image, and then the resulting segments are sent into a neural network that has already been trained. The constructed statistical classification-based Neural Network classifier has high accuracy, detecting and classifying these diseases that were put to the test in order with an accuracy rate of around 93%. This model has to be trained using a deep learning model, such as a convolutional neural network so that it can achieve higher levels of accuracy. Manimegalai and Sivakamasundari (2017) developed a system for automatic plant disease classification and identification using leaf image processing techniques. The model specifically focused on classifying apple leaves based on color and texture features, utilizing an SVM classifier. Color features were drawn out from color moments, and texture features were drawn out from the GLCM. Performance analysis revealed the accuracy of the model for texture, color, and combined features. The implementation of this technology can benefit farmers by improving selective application and productivity. However, the accuracy of detection can be improved by incorporating more features into the SVM classifier. Depending on the feature texture, Color, Texture, and color the accuracy was accordingly 97.22%, 96.32%, and 98.46%. Siburian et al. (2019) focused their attention on the interpretation of images for early-stage pest identification to protect the crop from being adversely affected. The cultivation of plants in greenhouses is the primary topic of investigation in this study. DBSCAN and NN computations are the techniques that are used here. According to the comprehensive dataset including pest information used in the study, the average precision of early pest detection is 96%. This proposed model gives the best result for only greenhouse crops which need more datasets for all environments.

Sivakamasundari and Seenivasagam (2018) proposed an algorithm based on SVM to test on apple leaves for detecting and identifying some particular diseases (Alternaria, Apple scab, and Cedar Rust) based on color feature and texture features. The classification accuracy based on texture was 92.22% and it was 92.67% based on color and both texture & color. In the future, the author wants to make their proposed system available for all kind of plant leaves and also display the name of the disease. Pavithra (2015) proposed an automated system for rice leaf disease detection using image processing techniques. The system employed feature extraction and image segmentation using SIFT, and an SVM classifier. The study primarily focused on Paddy Blast and brown spot diseases. The system aimed to improve the overall accuracy by 95.00%. The proposed system can facilitate plant leaf disease diagnosis without the need for expertise, assisting farmers in assessing their crops accurately. Early disease detection enables timely pest control techniques, minimizing risks to humans and the environment. The classifier SVM has an accuracy of 95.5% in the case of Blast 30 and Brow Spot 30. And the classifier KNN has an accuracy of 92.2% in the case of Blast 30 and Brow Spot 30. Akhtar et al. (2013) made a comparison among five different machine learning techniques: KNN, Naïve Bayes Classifier, SVM, Decision Tree, and Recurrent Neural Networks for the recognition and classification of crop diseases. Rose leaf samples were used as datasets which were acquired from a Tea Research Institute, which is located in Mansehra, and were separated into normal and diseased subsets. The subset which was labeled as diseased contains samples of Anthracnose and Black

Spots diseases. An incorporation of features DWT, DCT, and classifier SVM displayed the most accuracy in the result which is 94.45%. However, Decision Tree performed as the best classifier with an accuracy of 91.95% on the feature DCT when an individual feature was taken into consideration. Shrestha et al. (2020) have presented a technique that helps in determining the cause of plant diseases and provides remedies that may be used as a means of disease prevention. The different parts of the Internet-accessible database are set up correctly, and the different databases are kept separate. To make a good database, plant species are found and given new names. Then, a test database of different plant diseases is made to make sure the project is accurate and reliable. Plant disease can be found by looking at where it is on the leaves of a sick plant. The method used to find plant diseases is picture processing with Convolutional Neural Network (CNN). The suggested system has been developed in Python and achieves an accuracy of 78%. The processing speed and precision may be enhanced by using Google's GPU. PrajwalGowda et al. (2017) created a Machine Learning model adopting the Convolutional Neural Network (CNN) method in order to identify diseases in rice crops from images of their leaves. The researchers created a dataset of paddy leaves from both infected and uninfected rice plants. The CNN algorithm's model-training and disease-detection stages relied on this dataset for training. The proposed system offers a robust and cost-efficient solution, outperforming existing methods. Future work can involve training the system for other paddy crop diseases and developing user-friendly mobile applications.

Tulshan and Raul (2019) proposed KNN classifier which was compared against Linear SVM classifier for the analysis of various techniques for plant leaf disease detection utilizing the technique of image processing as well as to improve the current system. As a result, KNN performs more accurately and efficiently than SVM because Linear SVM can only categorize the data into two classes since it is a multi-class classification. So as a consequence, when the number of diseases is more than two, it displays inaccuracy. But on the contrary, the KNN classifier not only can determine the number of diseases but also conveys the area of a leaf (a leaf can be affected by multiple diseases at a time) that is defective by a particular disease. The accuracy of the suggested algorithm's (KNN) is 98.56% whereas the current system (Linear SVM) has an accuracy of 97.6%. Diseases that are addressed in this paper are White Fly, Leaf Miner, Mosaic Virus, Early Blight, and Down Mildew. The dataset consisted of 75 images. Geetharamani and Pandian (2019) have proposed a method for the identification of plant leaf diseases and offer solutions that can be used as a disease protection mechanism. They trained several models using their plant village dataset for identifying disease. These algorithms include Logistic Regression, K-Nearest Neighbors, Support vector machine, and Convolutional neural network. The suggested Deep CNN model accurately distinguishes between 38 types of healthy and ill plants based on photographs of their leaves. The Deep CNN model that was suggested produces superior prediction accuracy than other models, which vary from 50% to 87% of the time. The Deep CNN model's accuracy is 96.46%. With some fine-tuning methods, the improved information will make the model work better and be more accurate. Harini and Savitha (2019) focused on potato, rice, and bell pepper crops, analyzing various diseases for each. A Graphical User Interface (GUI) was developed for easy user interaction. The GUI allows users to browse files, upload images, view results, and exit. The project has the potential for further expansion, providing suggestions and remedies for diseases detected, and supporting farmers in disease management.

## Table 1. Comparative Study of Disease Detection in various plant leaves

| Author | Year | Techniques (Algorithm) | Detected Plants | Outcome |
|---|---|---|---|---|
| Fenu and Malloci (2019) | 2019 | ANN, SVM | Potato | The prediction accuracy for ANN was 96% and for SVM Classification was 98% |
| Ahmad et al. (2023) | 2022 | Deep learning | Multiple crops | An autonomous end-to-end plant disease control system may benefit from this sophisticated plant disease assessment by better detecting crops, and identifying their particular diseases early in the growing season, and precisely evaluating disease levels. |
| Ramesh et al. (2018) | 2018 | Logistic Regression, Random Forest, KNN, SVM, CART, Naive Bayes | Multiple crops | The Random forest Classification achieved the highest performance with an accuracy of 70%. |
| D. Al Bashish et al. (2010) | 2010 | Convolutional Neural Network (CNN) | Multiple crops | The constructed statistical classification-based Neural Network classifier has high accuracy, detecting and classifying these diseases that were put to test in order with an accuracy rate of around 93% |
| Manimegalai and Sivakamasundari (2017) | 2017 | SVM | Multiple crops | Accuracy depending upon the features (texture, color, texture, and color) were accordingly 97.22, 96.32, 98.46 |
| Siburian et al. (2019) | 2019 | DBSCAN, NN computations | Greenhouse crops | The average precision of early pest detection is 96% |
| Sivakamasundari and Seenivasagam (2018) | 2019 | SVM | Apple leaves | The classification accuracy based on texture was 92.22% and it was 92.67% based on color and both texture & color |
| Pavithra (2015) | 2015 | SIFT, SVM | Rice leaves | The overall accuracy of the system should improve by 95.00 percent. |
| A. Akhtar et al. (2013) | 2013 | KNN, Naïve Bayes, SVM, Decision Tree, Recurrent Neural Networks | Rose leaves | The combined approach of DCT, DWT, and the SVM classifier yields the highest accuracy (94.45%). However, when it comes to feature isolation, Decision Tree emerges as the most accurate classifier, with an accuracy of 91.95 on the DCT feature. |
| Shrestha et al. (2020) | 2019 | Convolutional Neural Network (CNN) | Multiple crops | The suggested system has been developed in Python and achieves an accuracy of 78% |
| PrajwalGowda et al. (2017) | 2020 | Convolutional Neural Network (CNN) | Rice leaves | Specifically, the technique relies on two steps: training the model and recognizing the supplied image, in order to detect the blast of rice and the bacterial blight of the leaves of paddy. |
| Tulshan and Raul (2019) | 2019 | KNN, Linear SVM | Different plant leaves | The proposed algorithm(KNN) accuracy is 98.56% while existing system(Linear SVM) has 97.6% of accuracy. |
| Geetharamani and Pandian (2019) | 2019 | Logistic Regression, KNN, SVM, CNN. | 14 Different plants | The Deep CNN model's accuracy is 96.46% |
| Harini and Savitha (2019) | 2021 | Convolutional Neural Network (CNN) | Potato, Rice, bell pepper | In this context, a graphical user interface (also known as a GUI) refers to a kind of user interface that enables users to interact with technological devices by means of graphical icons and components. |
| L Ahmed et al. (2019) | 2019 | KNN, Decision tree, Logistic regression, Naive Bayes | Rice leaves | Decision Tree performed the best on the test set with an accuracy of 97.9167%. |

**Source**: Existing Research papers

In (Ahmed et al. ,2019) four classification techniques were enforced to identify the diseases in rice leaves. The four algorithms that were compared were Naive Bayes, Logistic regression, Decision tree, and KNN. Among them, the Decision tree showed the best result on the test set with an accuracy of 97.9167%. In addition to accuracy, there were other measurements of performance which include AUC, recall value (Sensitivity), FPR, F-Measure, Precision value, and TPR which were also assessed to draw a comparison among the four algorithms and it was crystal clear that decision tree algorithm surpasses all other algorithms in every instance of disease classification and detection.

**Proposed Methodology**
The method that we used to conduct this research is quantitative. The following are the procedures we have used while carrying out the proposed model:

**1. Dataset Collection**
The plant images that we need to fulfill our requirements for different disease classes and healthy leaf images have been found in the plant village dataset in Kaggle (Plant Village., 2020), which is an online repository. These plant images had gathered with regard to bell pepper, tomato, and potato crops.

**2. Data Preprocessing**
The gathered images must be processed, cleaned, and transformed into a format that is simple to use. The diseases of crops are infected with are then properly labeled and marked on them (Chowdhury et al., 2023). We completed prior to fitting the raw data into the model, pre-processing was first accomplished. We used image pre-processing methods will help improve certain image features. To make the training robust, images are normalized and different augmentations are used.

**3. Feature Extraction**
Before training, the majority of traditional machine learning algorithms begin by manually extracting characteristics from the data being used. On the other end of the spectrum, contemporary techniques of deep learning may be taught on image data in a simple way. During training, the deep neural network is provided with the ability to automatically learn how to extract helpful discriminative characteristics and does so on an as-needed basis.

**4. Model Training**
In this work, we used the CNN network for training in plant disease detection for efficient and better performance. We see that the previous work above gets better performance for some algorithms, and there is one of the convolutional neural networks. An input layer, many hidden layers, and an output layer make up a convolutional neural network. There are four distinct types of layers in the hidden layer: the Convolution layer, the refined Linear Unit, the pooling layer, and the fully connected layer. The proposed model's CNN structure is presented in Figure 1. The remaining levels are explained in further detail below:

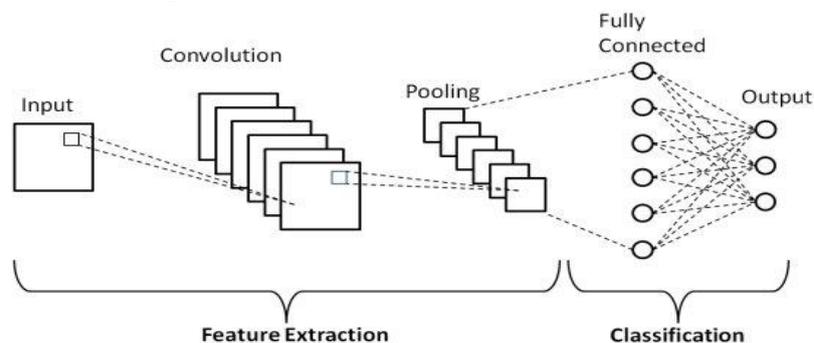

**Figure 1: Convolutional Neural Network Architecture (**Rhee ,2018)

**4.1) Convolutional Layer**: The convolutional layers transform a wide range of low-level data into higher-level characteristics that may be used for classification. Convolutional layers are also crucial to the operation of a Deep CNN. Initially, the Convolutional layers collect features from the training data, and the pooling layers decrease the dimensionality of the data (Uchida et al. ,Uchida).

**4.2) Rectified Linear Unit (ReLu)**: As one of many activation functions used in what is often referred to as an "activation function layer," ReLu is often mentioned. The ReLu activation function is employed in the hidden layers of our model. It is the triggering function that is used the most. In the ReLu layer, pixels with negative values are replaced with pixel values of 0, and the rest of the pixels stay the same (Wang et al. ,2020).

**4.3) Pooling Layer**: The process of downsizing in the spatial dimension is performed via the pooling layer. It allows for fewer parameters to be used. The Pooling layer shrinks the convolutional Feature in space, much like the Convolutional layer. Dimensionality reduction helps regulate overfitting and the amount of computing power needed to handle the data. This process uses a window to scan an image and then compresses the features used to extract the image. The most prevalent pooling layer approaches are average pooling and max pooling. Maximum pooling takes the greatest pixel value inside the image's chosen window, whereas average pooling takes an average of all pixel values (Gholamalinezhad and Khosravi ,2020). In Our model, we used 2*2 max pooling for training and get better precision for this pooling.

**4.4) Fully Connected Layer**: In order to convert an image into a vector, we stack convolution, RELU, and pooling layers multiple times after the first convolution. Classification work actual will be carried out in this layer. All of the neurons in this layer are linked to one another, and together they generate an N-dimensional vector containing the characteristics derived from the image. The last few layers of the network are composed of fully connected layers. The last layer in the process of predicting the class of an image is referred to as the dense layer or the fully connected layer (Basha et al. ,2020). The model is trained using the training dataset, and then the results are assessed using the test dataset and compared to the target value.

**5. Classification and Recognition:**
The image of leaves from a plant is taken as input by the model, and it is then analyzed for use in subsequent classification tasks. The trained CNN model is intended to be able to distinguish healthy and unhealthy leaves, as well as recognize diseases associated with the unhealthy input image when validation data is used during the testing phase. The CNN model is able to classify and identify both the disease that has affected the plant as well as the plant's name.

**Result and Discussion**
In agriculture, the growth of crops with diseases will reduce the quality and productivity of the harvest, which will have a negative impact on the crop's financial value; thus, the right deep learning techniques may be employed to recognize and classify all of the diseases impacted. We have collected data on three different kinds of crops: bell peppers, tomatoes, and potatoes. In the case of bell peppers, both bacterial spots and health are present. There are three types of potatoes: early blight, late blight, and healthy. Diseases that may affect tomatoes include early blight, late blight, bacterial spot, leaf mold, septoria leaf spot, spider mites two-spotted, target spot, mosaic virus, and healthy. The plant leaf disease dataset was used for both the training and the testing of the proposed CNN model. The dataset had been separated into training and testing sets, which had 17430 images. These images were labeled with 14 separate classes of damaged and healthy plant leaves, as well as background images. A comparison was made between the suggested model and the SVM, logistic regression, decision tree, and K-NN models. In addition, the following testing processes are evaluated and compared with regard to the results produced by the models. In the final analysis, the findings demonstrate that the proposed model outperforms the aforementioned alternatives. The proportion of right predictions that a model generates in comparison to the whole amount of predictions that are generated is referred to as its accuracy. The prediction success rate for the proposed Deep CNN model is 85.31%, which is higher than the 41%-67% achieved by alternative approaches.

**Table 2. Accuracy on Training and Test Dataset**

| Algorithms | Accuracy on the Training set | Accuracy on the Testing set |
|---|---|---|
| SVM | 82.93% | 66.15% |
| Logistic Regression | 49.01% | 42.43% |
| KNN | 51.82% | 41.25% |
| CNN | 99% | 85.31% |

Source: Coding part using dataset

According to the proposed CNN Method, When training a deep learning model, an epoch is a hyperparameter that specifies one whole iteration through the training data. As shown in Fig. 2(b), the suggested CNN model with 150 epochs employing the enhanced dataset attained an accuracy for validation of 85.31%. Another hyperparameter is a batch size in which accuracy was increased with different batch size values, shown in Fig. 2(c).

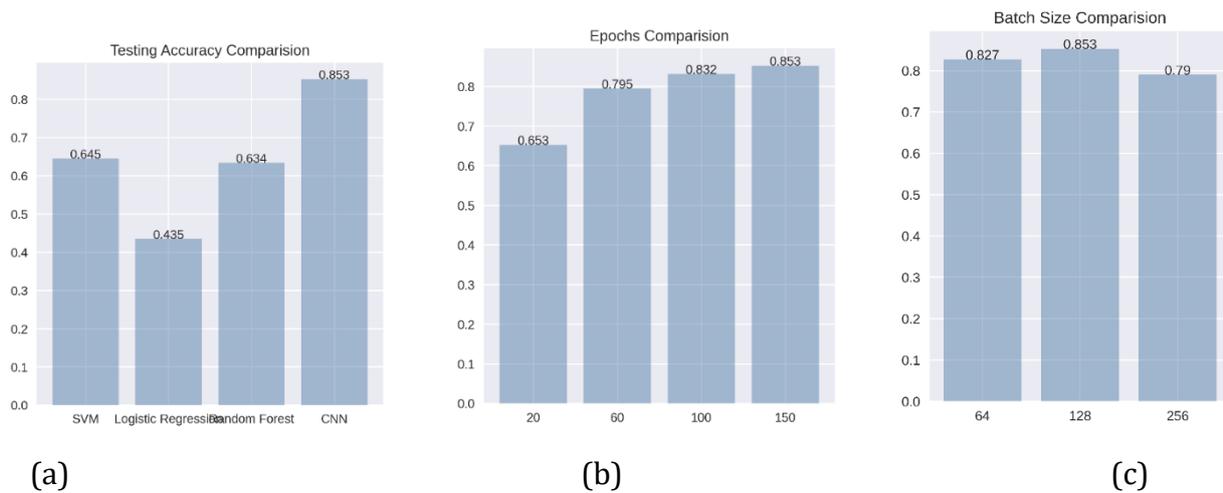

(a)                                (b)                                (c)

**Figure 2:** (a) Testing Accuracy of the different Models, (b) Testing Accuracy of the different Epochs, (c) Testing Accuracy of the different Batch Sizes

**Conclusion**

It's apparent that CNN has efficiently improvised profoundly in detecting plant diseases. It's been proven profitable for farmers abroad. Hence, we proposed a better CNN model for detecting the diseases of plants that are mostly produced in Bangladesh. A comparison has been made between different machine learning algorithms and among them, the best outcome was derived from CNN. Consequently, we proposed a more excelling CNN model which has given an accuracy of 85.31%. Our model has been trained with the online data sets which are mostly collected from other countries. The majority of the research used plants from other countries. This has fewer advantages for farmers in Bangladesh. Therefore, our primary goal was to provide a result that would help Bangladeshi farmers. Consequently, we worked with plants that are mostly grown in Bangladesh. Also later on, we'll use pre-trained transfer learning model for more accurate performance instead of machine learning.

**Limitations and Future Research**
We've worked on online data sets. The lack of availability has prevented us from working with any Bangladeshi data sets. Therefore, we want to gather our own data from Bangladeshi fields in the future so that we may train our model from scratch. We used a regular CPU to work with. We plan to use highly configured GPUs in the future, which should further improve overall accuracy. We worked with online datasets of plants, the majority of which are grown in Bangladesh. So far, we've worked with only three types of plants- bell pepper, tomato, and potato. As rice is our staple food so the data on rice plants will also be included for the identification of rice leaf diseases. In addition to this, we would use a highly configured GPU which will expectantly increase the overall accuracy even more. Furthermore, we'll develop a mobile application based on our model which will also be accessible offline. Since the internet is not that prevalent in rural areas of Bangladesh, this application will mostly be beneficial to the farmers.

**Funding**
*Authors did not receive any funding or grants for this project.*

**Conflict of interest**
*The authors claim that there were no competing interests in this work. All of the authors of this article have given their consent for it to be published.*

**Cite this article:**

# Published by

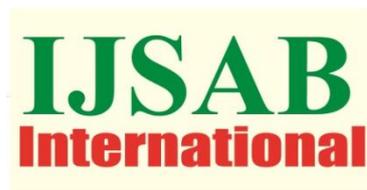
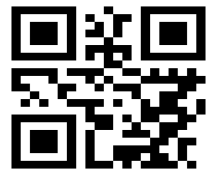